# HQSI：混合量子群體智慧─
# 以線上憑證狀態協定請求流量預測為例

Abel C. H. Chen
*Information & Communications Security Laboratory,
Chunghwa Telecom Laboratories*
Taoyuan, Taiwan
ORCID: 0000-0003-3628-3033

*摘要*—隨著量子計算技術日益成熟，產官學研各界也開始著重量子計算未來的應用。搭配人工智慧技術的方法，多種量子神經網路(Quantum Neural Network, QNN)模型也陸續被提出，包含量子卷積神經網路(Quantum Convolutional Neural Network, QCNN)、量子長短期記憶(Quantum Long Short-Term Memory, QLSTM)網路、量子生成對抗網路(Quantum Generative Adversarial Network, QGAN)等，並且在最佳化方法則有線性近似約束最佳化(Constrained Optimization By Linear Approximation, COBYLA)、同時擾動隨機近似(Simultaneous Perturbation Stochastic Approximation, SPSA)。有鑑於此，本研究提出混合量子群體智慧(Hybrid Quantum Swarm Intelligence, HQSI)，通過建構量子神經網路模型作為前向傳播神經網路(Forward Propagation Neural Network, FPNN)，量測量子態和預測結果後代入經典電腦的群體智慧演算法進行權重最佳化，訓練期間在量子電腦和經典電腦之間迭代完成。在實驗階段，本研究採用線上憑證狀態協定請求流量預測為例，驗證本研究提出的混合量子群體智慧對比最先進的(State-of-the-Art, SOTA)量子最佳化演算法，從實驗結果顯示本研究提出的混合量子群體智慧可以降低一半以上的誤差。

*關鍵字—群體智慧、量子神經網路、量子計算、請求流量預測、線上憑證狀態協定*

## I. 前言

近年來，人工智慧非常熱門，每年都有許多模型推陳出新，也帶動了各種生活中的應用與服務[1]-[2]。並且，隨著量子計算技術的成熟，許多研究也開始探索量子人工智慧領域，開始提出各種量子神經網路(Quantum Neural Network, QNN)模型[3]-[5]也陸續被提出，包含量子卷積神經網路(Quantum Convolutional Neural Network, QCNN)[6]-[7]、量子長短期記憶(Quantum Long Short-Term Memory, QLSTM)網路[8]-[9]、量子生成對抗網路(Quantum Generative Adversarial Network, QGAN)[10]等，嘗試用量子疊加態和量子糾纏態等特性來實現指數級加速，同時在量子計算上可能取得經典電腦神經網路所無法提取的特徵，進而提升模型的精確度。此外，為有效訓練量子神經網路，許多量子最佳化演算法也陸續被提出，包含線性近似約束最佳化(Constrained Optimization By Linear Approximation, COBYLA)[11]-[12]、同時擾動隨機近似 (Simultaneous Perturbation Stochastic Approximation, SPSA)[13]等，可以用較高的效率來訓練量子神經網路。

有鑑於此，本研究提出混合量子群體智慧(Hybrid Quantum Swarm Intelligence, HQSI)，在量子電腦中建構量子神經網路模型，並且該量子神經網路是一種前向傳播神經網路(Forward Propagation Neural Network, FPNN)，可以把輸入變量和權重變量代入後進行量子邏輯閘計算，以及量測量子態後得到結果。之後再把量測結果代入經典電腦群體智慧演算法的目標函數中，根據目標函數進行權重變量最佳化。訓練過程中，在量子電腦執行量子神經網路計算和量測，以及在經典電腦執行群體智慧演算法和權重變量最佳化，不斷在量子電腦和經典電腦之間迭代。當訓練完成後，最佳化後的權重變量值即可作為後續在運行(run-time)階段量子神經網路的參數值，並且提供實際預測使用。本研究的主要貢獻條列如下：

1. 本研究提出混合量子群體智慧，可以結合經典電腦群體智慧演算法和量子電腦量子神經網路。並且，本研究方法屬於一種範式，未來可以適用任何一種群體智慧演算法和任何一種量子神經網路。

2. 本研究採用線上憑證狀態協定請求流量預測為例進行實證，通過實驗結果證實在本研究案例中，採用混合量子群體智慧可以比最先進的(State-of-the-Art, SOTA)量子最佳化演算法提供較低的預測誤差。

3. 本研究採用的群體智慧演算法包含有粒子群最佳化(Particle Swarm Optimization, PSO)演算法[14]-[16]和基因演算法(Genetic Algorithm, GA)[17]-[19]，可以作為一種展示，讓讀者可以通過本研究提出的範式來發展出自己的應用。

本論文主要分為 5 節。第 II 節描述量子神經網路、粒子群最佳化演算法、以及基因演算法。第 III 節描述本研究提出的混合量子群體智慧之主要構想，並且在第 IV 節驗證本研究提出的混合量子群體智慧之效能。最後，第 V 節總結本研究貢獻，並且討論未來可行的研究方向。

## II. 相關研究

本節將分別介紹量子神經網路、粒子群最佳化演算法、以及基因演算法的基本構想。

*A. 量子神經網路*

本節首先介紹量子神經網路會用到的量子邏輯閘，然後再介紹量子神經網路的結構。

*1) 量子態與量子邏輯閘*

作者在自己先前的研究[4]已經詳細介紹一個量子位元時的簡單量子神經網路(Simple Quantum Neural Network)，有興趣的讀者可以從該文獻中找到相關資訊。常用到的量子邏輯閘包含有 Hadamard Gate 和 RY Gate，通過 Hadamard Gate 建立量子均勻疊加態，並且用 RY Gate 可以用來代入輸入變量集合 $X$ 和權重變量集合 $W$，作為對量子位元 Y 軸旋轉的角度值。除此之外，當如果有需要用到兩個以上的量子位元時，則可以加入 CNOT

Gate 來建立量子糾纏態,用以提取多個變量之間交互影響後的特徵值。表 I 為 Hadamard Gate、RY Gate、以及 CNOT Gate 的圖示及其數學表示式。當兩個量子位元 $q_i$、$q_j$分別操作 Hadamard Gate,其量子態可以用公式(1)表示。而對兩個量子位元 $q_i$、$q_j$分別操作 RY Gate 旋轉$\theta_i$、$\theta_j$度角,其量子態可以用公式(2)表示,詳細推導可參考附錄 A。

$$H \otimes H |q_i q_j\rangle \tag{1}$$

$$R(\theta_i) \otimes R(\theta_j)|q_i q_j\rangle \tag{2}$$

TABLE I. QUANTUM LOGIC GATES

| Logic Gate | Notation | Symbol |
|---|---|---|
| Hadamard Gate | $H\|q_j\rangle = \frac{1}{\sqrt{2}}\begin{bmatrix}1 & 1\\1 & -1\end{bmatrix}\|q_j\rangle$ |  |
| RY Gate | $R(\theta)\|q_j\rangle$ $= \begin{bmatrix}cos\left(\frac{\theta}{2}\right) & -sin\left(\frac{\theta}{2}\right)\\sin\left(\frac{\theta}{2}\right) & cos\left(\frac{\theta}{2}\right)\end{bmatrix}\|q_j\rangle$ |  |
| CNOT Gate Control Qubit: $q_j$ Target Qubit: $q_i$ | $C(q_j)\|q_i q_j\rangle$ $= \begin{bmatrix}1 & 0 & 0 & 0\\0 & 0 & 0 & 1\\0 & 0 & 1 & 0\\0 & 1 & 0 & 0\end{bmatrix}\|q_i q_j\rangle$ | 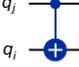 |

### 2) 量子神經網路結構

量子神經網路主要包含輸入層(即特徵圖)、隱藏層(即變分模型)、輸出層(即量測),以兩個量子位元為例的結構圖如圖 1 所示。

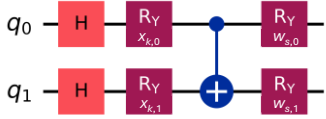

Fig. 1. The structure of QNN for two qubits.

本研究假設$|q_1 q_0\rangle$初始量子態為$|00\rangle$。其中,輸入層主要用來設定輸入變量,在此例中每筆資料有兩個特徵值,第 $k$ 筆資料特徵值分別表示為 $x_{k,0}$ 和 $x_{k,1}$ 兩個輸入變量,可以通過操作 RY Gate 輸入到量子神經網路,如公式(3)表示。隱藏層主要操作 CNOT Gate 產生量子糾纏態,用以提取出變量交互影響後的特徵,以及通過操作 RY Gate 輸入第 s 組解的兩個權重 $w_{s,0}$ 和 $w_{s,1}$ 到量子神經網路,如公式(4)表示。輸出層則是量測$|q_1 q_0\rangle''$量子態,並且根據量測結果代入目標函數產生目標值;例如:迴歸應用的目標函數可以是平均平方誤差(Mean Squared Error, MSE),分類應用的目標函數可以是交叉熵(cross-entropy)。為有效擬合真值,可以加入多層隱藏層,通過更多的權重變量來調整,以加第 $T$ 層隱藏層為例,其範式如公式(5)所示,詳細推導可參考附錄 A。

$$\left(R(x_{k,1}) \otimes R(x_{k,0})\right)(H \otimes H)|q_1 q_0\rangle = |q_1 q_0\rangle' \tag{3}$$

$$\left(R(w_{s,1}) \otimes R(w_{s,0})\right)C(q_0)|q_1 q_0\rangle' = |q_1 q_0\rangle'' \tag{4}$$

$$\left(\prod_{t=1}^{T}\left(R(w_{s,2(T-t)+1}) \otimes R(w_{s,2(T-t)})\right)C(q_0)\right)|q_1 q_0\rangle' \tag{5}$$

### B. 粒子群最佳化演算法

粒子群最佳化演算法在最佳化的過程中,本研究主要根據目標函數值$F(X, W)$計算最小值作為最佳化目標。其中,假設粒子數量總共有 $M$ 個,每個粒子有 $L$ 個特徵值,第 $s$ 個粒子在第 $r$ 個回合時的個體最佳解為$P_s^{(r)}$ (如公式(6)所示),在第 $r$ 個回合時的群體最佳解為$G^{(r)}$ (如公式(7)所示)。在粒子群最佳化演算法中包含第 $s$ 個粒子第 $l$ 個特徵在第 $r$ 個回合時的慣性速度為$v_{s,l}^{(r)}$、粒子變量第 $l$ 個特徵與目前個體最佳解的距離為$(p_{s,l}^{(r)} - w_{s,l}^{(r)})$、粒子變量第 $l$ 個特徵與目前群體最佳解的距離為$(g_l^{(r)} - w_{s,l}^{(r)})$來最佳化粒子變量第 $l$ 個特徵值,如公式(8)所示[15]。因此,需要設定超參數慣性速度權重$\alpha$、個體最佳解權重$c_1$、群體最佳解權重$c_2$,用以調整粒子變量第 $l$ 個特徵值,如公式(9)所示[15]。其中,$\beta_1$和$\beta_2$為隨機數。

$$P_s^{(r)} = \underset{0 \leq \zeta \leq r}{\text{argmin}} F\left(X, W_s^{(\zeta)}\right)$$
$$= \{p_{s,0}^{(r)}, p_{s,1}^{(r)}, \ldots, p_{s,L-1}^{(r)}\} \tag{6}$$

$$G^{(r)} = \underset{0 \leq s \leq M-1}{\text{argmin}} F\left(X, P_s^{(r)}\right)$$
$$= \{g_0^{(r)}, g_1^{(r)}, \ldots, g_{L-1}^{(r)}\} \tag{7}$$

$$v_{s,l}^{(r+1)} = \alpha v_{s,l}^{(r)} + \beta_1 c_1\left(p_{s,l}^{(r)} - w_{s,l}^{(r)}\right) + \beta_2 c_2\left(g_l^{(r)} - w_{s,l}^{(r)}\right) \tag{8}$$

$$w_{s,l}^{(r+1)} = w_{s,l}^{(r)} + v_{s,l}^{(r+1)} \tag{9}$$

### C. 基因演算法

基因演算法在最佳化的過程中主要包含選擇、交配、突變三個運算,本研究主要根據目標函數值$F(X, W)$計算最小值作為最佳化目標,流程圖如圖 2 所示。其中,假設基因序列數量總共有 $M$ 個,每個基因序列有 $L$ 個特徵值;在初始階段隨機產生每個基因序列的特徵值,並且計算對應的目標函數值。在選擇運算時,第 $s$ 個基因序列在第 $r$ 個回合對應的目標函數值$F(X, W_s^{(r-1)})$將決定第 $s$ 個基因序列在第 $r$ 個回合被選擇到的機率$\varphi_s^{(r)}$,如公式(10)所示;當目標函數值越小,被選擇的機率越高。之後再根據超參數交配率$\varsigma_1$決定有多少基因序列參與交配,以及隨機建立交配點來交換部分基因序列內容。以及根據超參數突變率$\varsigma_2$來決定基因序列是否突變,以及對參與突變的基因序列隨機對部分內容進行突變運算[18]。最後,再把經過選擇、交配、突變運算後的 $M$ 個基因序列代入目標函數,取得每組基因序列對應的目標函數值,並且紀錄目標函數值最小的基因序列(即該回合的最佳解),再進行迭代直到收斂。

$$\varphi_s^{(r)} = \frac{\rho_s^{(r)}}{\sum_{\xi=0}^{M-1} \rho_\xi^{(r)}}, \text{where } \rho_s^{(r)} = \frac{1}{F(X, W_s^{(r-1)})} \tag{10}$$

### III. 混合量子群體智慧

本節將先介紹本研究提出的混合量子群體智慧的主要流程,再介紹本研究提出的線上憑證狀態協定請求流量預測的量子神經網路結構。最後再描述本研究提出的群體智慧演算法的目標函數。

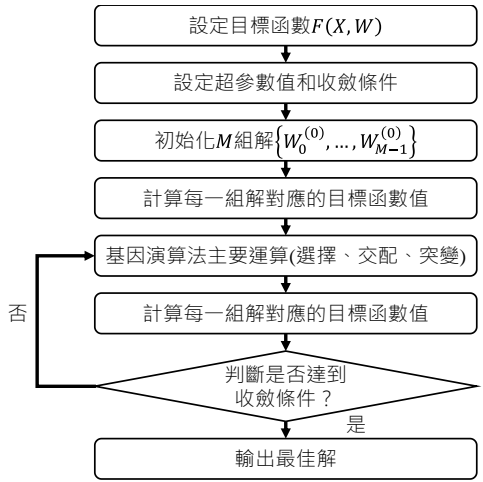

Fig. 2. The procedure of GA.

### A. 主要流程

本研究提出的混合量子群體智慧方法主要是在經典電腦上執行群體智慧最佳化演算法,以及在量子電腦上執行量子神經網路計算,整體流程如圖 3 所示。

初始階段:在量子電腦先根據要解決的應用問題建立量子神經網路,具體結構於下一個小節中說明。在經典電腦執行群體智慧最佳化演算法的初始化,包含設定目標函數、超參數、收斂條件,以及初始化 $M$ 組解。之後再把資料集內容作為輸入變量 $X$ 和每一組解作為權重變量 $W$ (例如:第 $s$ 組解為 $W_s^{(0)}$) 傳送到量子電腦。量子電腦把輸入變量 $X$ 和權重變量 $W$ 作為 RY Gate 角度值代入量子神經網路,並且執行量子神經網路,以及量測結果 $|q_5q_4q_3q_2q_1q_0\rangle''$。之後再由量子電腦傳送量測結果 $|q_5q_4q_3q_2q_1q_0\rangle''$ 給經典電腦,由經典電腦計算每一組解對應的目標函數值。

訓練階段:在訓練過程,將由經典電腦執行群體智慧演算法主要運算更新每一組解。其中,粒子群最佳化演算法主要運算包含根據前一回合的個體最佳解和群體最佳解來更新速度和更新權重。基因演算法主要運算包含選擇、交配、突變,根據前一回合每組解對應的目標函數值讓相對較佳解可以有更高的機率被選上和參與交配和突變。在更新完每一組解後,再把把資料集內容作為輸入變量 $X$ 和每一組解作為權重變量 $W$ (例如:第 $\zeta$ 回合第 $s$ 組解為 $W_s^{(\zeta)}$) 傳送到量子電腦,由量子電腦執行量子神經網路和量測結果。之後再由經典電腦計算每一組解對應的目標函數值和判斷是否達到收斂條件。本研究的收斂條為回合數上限 100 或損失值與前一回合的差異低於 $10^{-8}$。當達到收斂條件時,則輸出當下找到的最佳解。

### B. 量子神經網路結構

本研究解決的應用問題是線上憑證狀態協定請求流量預測,主要輸入變量是前一週的請求流量,輸出變量是後一週的請求流量,可以根據預測的請求流量來產生請求預簽章,避免流量高峰和達到流量分流效果,詳細應用可以參考作者先前研究[20]。

為解決此應用問題,並且由於週日的請求流量較少,所以本研究主要分析週一到週六的請求流量。因此,每筆輸入變量有 6 個特徵值,所以建構具有 6 個量子位元的量子神經網路結構,如圖 4 所示。然後分別操作 Hadamard Gate 產生均勻疊加態,再操作 RY Gate 帶入輸入變量。之後為兩兩量子位元操作 CNOT Gate 產生量子糾纏態,再操作 RY Gate 帶入權重變量,在此例中共有 6 個權重變量需要進行最佳化。

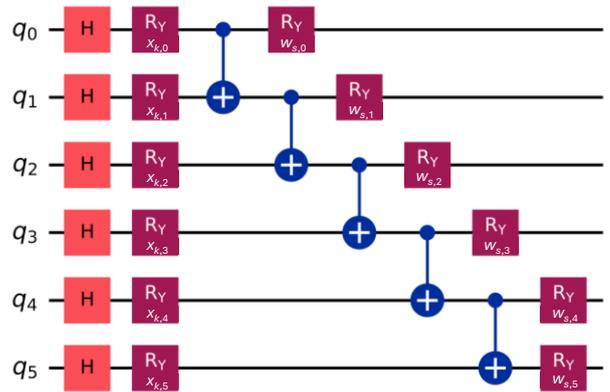

Fig. 4. The structure of QNN for OCSP request flow prediction.

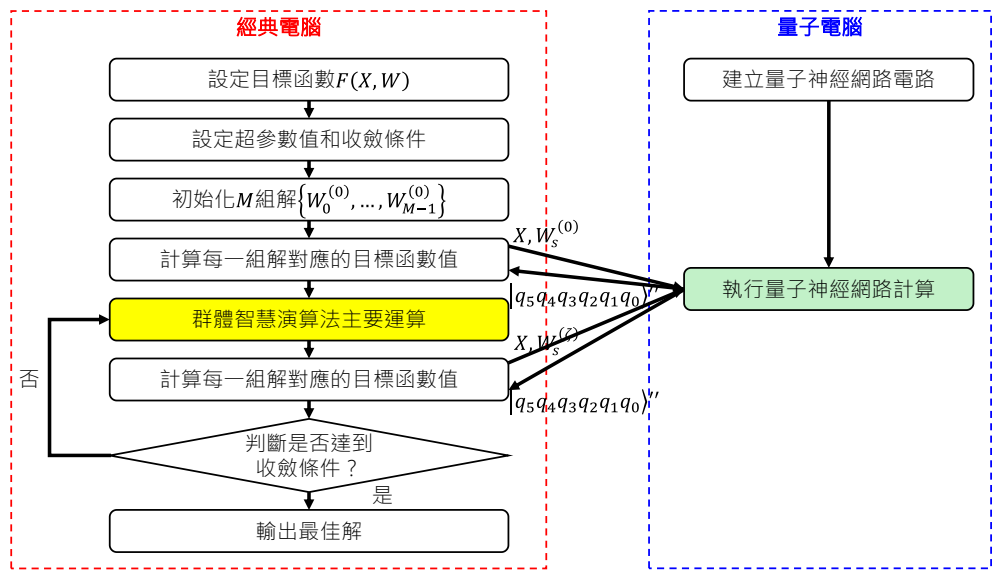

Fig. 3. The procedure of the proposed Hybrid Quantum Swarm Intelligence.

## C. 群體智慧演算法

本研究採用的群體智慧演算法主要包含粒子群最佳化演算法和基因演算法，在群體智慧演算法主要運算的部分主要延用原本的運算方法，所以本節主要著重於定義本研究提出的目標函數。由於量子神經網路量測結果總共有 6 個量子位元的量子態，在第 $k$ 筆第 $l$ 個量子位元$|q_l\rangle$屬於$|0\rangle$的機率是$v_{k,l,0}^2$，屬於$|1\rangle$的機率是$v_{k,l,1}^2$，可運用公式(11)得到估計值$y'_{k,l}$。之後再計算 $N$ 筆資料的平均平方誤差值為損失值(即目標函數值)。需要注意的是，由於目標函數值的值域是[–1, 1]，所以真值$y_{k,l}$在初始階段需正規化為同樣的值域範圍。

$$loss = \frac{1}{N}\sum_{k=1}^{N}\sum_{l=0}^{L-1}(y_{k,l} - y'_{k,l})^2, \quad (11)$$

$$\text{where } y'_{k,l} = v_{k,l,0}^2 - v_{k,l,1}^2$$

## IV. 實驗結果與討論

本節首先將介紹實驗環境及限制，然後先用簡單案例驗證本研究方法，再討論線上憑證狀態協定請求流量預測實驗結果，並且討論方法的優缺點。

### A. 實驗環境

本研究採用文獻[20]的資料集進行本研究方法的驗證，並且在 Intel(R) Core(TM) i7-10510U CPU 和 16 GB RAM 的硬體環境執行 Qiskit 建立量子模擬器。採用的軟體規格如表 II 所示。此外，本研究採用的 $N$ 值為 20，收斂條件已於第 III.A 節中說明。

TABLE II. THE USED SOFTWARE IN EXPERIMENTS

| Model and Method | Package | Version |
|---|---|---|
| QNN | Qiskit | 1.1.1 |
| PSO | Pyswarm | 0.6 |
| GA | Deap | 1.4 |

### B. 簡單模型實驗結果比較

為充分驗證本研究提出的方法，本節採用簡單神經網路擬合線性函數、Sigmoid 函數、Tanh 函數(詳細設置可參考[4])，並與常用的量子最佳化演算法 COBYLA 和 SPSA 對比，實驗結果分別如表 III、表 IV、表 V 所示。由實驗結果可以觀察到，本研究提出的 HQSI 在損失值並不高於 COBYLA 和 SPSA，表示本研究方法可以找到相同精度的解。唯需要注意的是，本研究方法由於同時找 $M$ 組解，所以需要較多的計算時間。

TABLE III. COMPARISON RESULTS IN THE CASE OF LINEAR FUNCTION

| Method | Computation Time (ms) | Loss |
|---|---|---|
| The Proposed HQSI Based on PSO | 24814.1546 | 0.0006 |
| The Proposed HQSI Based on GA | 93382.9877 | 0.0006 |
| COBYLA [11]-[12] | 4113.8704 | 0.0006 |
| SPSA [13] | 94687.4443 | 1.0946 |

TABLE IV. COMPARISON RESULTS IN THE CASE OF SIGMOID FUNCTION

| Method | Computation Time (ms) | Loss |
|---|---|---|
| The Proposed HQSI Based on PSO | 24681.0060 | 0.0003 |
| The Proposed HQSI Based on GA | 93145.9450 | 0.0003 |
| COBYLA [11]-[12] | 3476.3585 | 0.0003 |
| SPSA [13] | 88251.2764 | 0.0003 |

TABLE V. COMPARISON RESULTS IN THE CASE OF TANH FUNCTION

| Method | Computation Time (ms) | Loss |
|---|---|---|
| The Proposed HQSI Based on PSO | 24573.0932 | 0.0003 |
| The Proposed HQSI Based on GA | 92855.0802 | 0.0003 |
| COBYLA [11]-[12] | 3552.2435 | 0.0003 |
| SPSA [13] | 89863.838 | 0.0003 |

### C. 線上憑證狀態協定請求流量預測模型實驗結果比較

表 IV 為線上憑證狀態協定請求流量預測的效能比較結果，可以觀察到本研究提出的方法可以達到最低的預測誤差率，僅約為 2%。COBYLA 和 SPSA 在此應用表現上，則預測誤差率約為 5.8%。然而，如同前述，本研究方法的缺點在於同時找 $M$ 組解，所以需要較多的計算時間。

TABLE VI. COMPARISON RESULTS IN THE CASE OF OCSP REQUEST FLOW PREDICTION

| Method | Computation Time (ms) | Prediction Error Rate |
|---|---|---|
| The Proposed HQSI Based on PSO | 30738.0451 | 2.0381% |
| The Proposed HQSI Based on GA | 41989.5371 | 1.9895% |
| COBYLA [11]-[12] | 10942.3984 | 5.8119% |
| SPSA [13] | 30386.3444 | 5.8121% |

## V. 結論與未來研究

本研究提出混合量子群體智慧，通過經典電腦執行群體智慧最佳化演算法，以及量子電腦執行量子神經網路計算，並且應用在線上憑證狀態協定請求流量預測，可以有效降低預測誤差率。

有鑑於本研究方法最大的限制在於計算時間較長，在未來研究中可以考慮把經典電腦上計算的部分運算設計為量子電路，改由量子電腦來計算和達到指數級加速。

## 附錄 A

本節主要提供詳細的量子邏輯閘對應的數學表示式推導，提供給讀者精確的表示式。兩個量子位元 $q_i$、$q_j$ 分別操作 Hadamard Gate 後，量子態的結果如公式(A1)所示。兩個量子位元 $q_i$、$q_j$ 分別操作 RY Gate 旋轉 $\theta_i$、$\theta_j$ 度角後，量子態的結果如公式(A2)所示。輸入層(即特徵圖)分別操作 RY Gate 旋轉 $x_i$、$x_j$ 度角後，量子態的結果如公式(A3)所示。隨後加入 1 層隱藏層(即變分模型，每一層隱藏層包含操作一個 CNOT Gate 和分別操作 RY Gate 旋轉 $w_1$、$w_0$ 度角) 後，量子態的結果如公式(A4)所示。而加入 $T$ 層隱藏層(即變分模型，每一層隱藏層包含操作一個 CNOT Gate 和分別操作 Ry Gate 旋轉) 後，量子態的結果如公式(A5)所示。

$$H \otimes H |q_i q_j\rangle = \frac{1}{\sqrt{2}}\begin{bmatrix}1 & 1 \\ 1 & -1\end{bmatrix} \otimes \frac{1}{\sqrt{2}}\begin{bmatrix}1 & 1 \\ 1 & -1\end{bmatrix}|q_i q_j\rangle = \frac{1}{2}\begin{bmatrix}1 & 1 & 1 & 1 \\ 1 & -1 & 1 & -1 \\ 1 & 1 & -1 & -1 \\ 1 & -1 & -1 & 1\end{bmatrix}|q_i q_j\rangle \tag{A1}$$

$$R(\theta_i) \otimes R(\theta_j)|q_i q_j\rangle = \begin{bmatrix}\cos\left(\frac{\theta_i}{2}\right) & -\sin\left(\frac{\theta_i}{2}\right) \\ \sin\left(\frac{\theta_i}{2}\right) & \cos\left(\frac{\theta_i}{2}\right)\end{bmatrix} \otimes \begin{bmatrix}\cos\left(\frac{\theta_j}{2}\right) & -\sin\left(\frac{\theta_j}{2}\right) \\ \sin\left(\frac{\theta_j}{2}\right) & \cos\left(\frac{\theta_j}{2}\right)\end{bmatrix}|q_i q_j\rangle$$

$$= \begin{bmatrix}\cos\left(\frac{\theta_i}{2}\right)\begin{bmatrix}\cos\left(\frac{\theta_j}{2}\right) & -\sin\left(\frac{\theta_j}{2}\right) \\ \sin\left(\frac{\theta_j}{2}\right) & \cos\left(\frac{\theta_j}{2}\right)\end{bmatrix} & -\sin\left(\frac{\theta_i}{2}\right)\begin{bmatrix}\cos\left(\frac{\theta_j}{2}\right) & -\sin\left(\frac{\theta_j}{2}\right) \\ \sin\left(\frac{\theta_j}{2}\right) & \cos\left(\frac{\theta_j}{2}\right)\end{bmatrix} \\ \sin\left(\frac{\theta_i}{2}\right)\begin{bmatrix}\cos\left(\frac{\theta_j}{2}\right) & -\sin\left(\frac{\theta_j}{2}\right) \\ \sin\left(\frac{\theta_j}{2}\right) & \cos\left(\frac{\theta_j}{2}\right)\end{bmatrix} & \cos\left(\frac{\theta_i}{2}\right)\begin{bmatrix}\cos\left(\frac{\theta_j}{2}\right) & -\sin\left(\frac{\theta_j}{2}\right) \\ \sin\left(\frac{\theta_j}{2}\right) & \cos\left(\frac{\theta_j}{2}\right)\end{bmatrix}\end{bmatrix}|q_i q_j\rangle$$

$$= \begin{bmatrix}\cos\left(\frac{\theta_i}{2}\right)\cos\left(\frac{\theta_j}{2}\right) & -\cos\left(\frac{\theta_i}{2}\right)\sin\left(\frac{\theta_j}{2}\right) & -\sin\left(\frac{\theta_i}{2}\right)\cos\left(\frac{\theta_j}{2}\right) & \sin\left(\frac{\theta_i}{2}\right)\sin\left(\frac{\theta_j}{2}\right) \\ \cos\left(\frac{\theta_i}{2}\right)\sin\left(\frac{\theta_j}{2}\right) & \cos\left(\frac{\theta_i}{2}\right)\cos\left(\frac{\theta_j}{2}\right) & -\sin\left(\frac{\theta_i}{2}\right)\sin\left(\frac{\theta_j}{2}\right) & -\sin\left(\frac{\theta_i}{2}\right)\cos\left(\frac{\theta_j}{2}\right) \\ \sin\left(\frac{\theta_i}{2}\right)\cos\left(\frac{\theta_j}{2}\right) & -\sin\left(\frac{\theta_i}{2}\right)\sin\left(\frac{\theta_j}{2}\right) & \cos\left(\frac{\theta_i}{2}\right)\cos\left(\frac{\theta_j}{2}\right) & -\cos\left(\frac{\theta_i}{2}\right)\sin\left(\frac{\theta_j}{2}\right) \\ \sin\left(\frac{\theta_i}{2}\right)\sin\left(\frac{\theta_j}{2}\right) & \sin\left(\frac{\theta_i}{2}\right)\cos\left(\frac{\theta_j}{2}\right) & \cos\left(\frac{\theta_i}{2}\right)\sin\left(\frac{\theta_j}{2}\right) & \cos\left(\frac{\theta_i}{2}\right)\cos\left(\frac{\theta_j}{2}\right)\end{bmatrix}|q_i q_j\rangle \tag{A2}$$

$$|q_1 q_0\rangle' = \left(R(x_{k,1}) \otimes R(x_{k,0})\right)(H \otimes H)|q_1 q_0\rangle$$

$$= \begin{bmatrix}\cos\left(\frac{x_{k,1}}{2}\right)\cos\left(\frac{x_{k,0}}{2}\right) & -\cos\left(\frac{x_{k,1}}{2}\right)\sin\left(\frac{x_{k,0}}{2}\right) & -\sin\left(\frac{x_{k,1}}{2}\right)\cos\left(\frac{x_{k,0}}{2}\right) & \sin\left(\frac{x_{k,1}}{2}\right)\sin\left(\frac{x_{k,0}}{2}\right) \\ \cos\left(\frac{x_{k,1}}{2}\right)\sin\left(\frac{x_{k,0}}{2}\right) & \cos\left(\frac{x_{k,1}}{2}\right)\cos\left(\frac{x_{k,0}}{2}\right) & -\sin\left(\frac{x_{k,1}}{2}\right)\sin\left(\frac{x_{k,0}}{2}\right) & -\sin\left(\frac{x_{k,1}}{2}\right)\cos\left(\frac{x_{k,0}}{2}\right) \\ \sin\left(\frac{x_{k,1}}{2}\right)\cos\left(\frac{x_{k,0}}{2}\right) & -\sin\left(\frac{x_{k,1}}{2}\right)\sin\left(\frac{x_{k,0}}{2}\right) & \cos\left(\frac{x_{k,1}}{2}\right)\cos\left(\frac{x_{k,0}}{2}\right) & -\cos\left(\frac{x_{k,1}}{2}\right)\sin\left(\frac{x_{k,0}}{2}\right) \\ \sin\left(\frac{x_{k,1}}{2}\right)\sin\left(\frac{x_{k,0}}{2}\right) & \sin\left(\frac{x_{k,1}}{2}\right)\cos\left(\frac{x_{k,0}}{2}\right) & \cos\left(\frac{x_{k,1}}{2}\right)\sin\left(\frac{x_{k,0}}{2}\right) & \cos\left(\frac{x_{k,1}}{2}\right)\cos\left(\frac{x_{k,0}}{2}\right)\end{bmatrix}\frac{1}{2}\begin{bmatrix}1 & 1 & 1 & 1 \\ 1 & -1 & 1 & -1 \\ 1 & 1 & -1 & -1 \\ 1 & -1 & -1 & 1\end{bmatrix}\begin{bmatrix}1 \\ 0 \\ 0 \\ 0\end{bmatrix}$$

$$= \frac{1}{2}\begin{bmatrix}\cos\left(\frac{x_{k,1}}{2}\right)\cos\left(\frac{x_{k,0}}{2}\right) & -\cos\left(\frac{x_{k,1}}{2}\right)\sin\left(\frac{x_{k,0}}{2}\right) & -\sin\left(\frac{x_{k,1}}{2}\right)\cos\left(\frac{x_{k,0}}{2}\right) & \sin\left(\frac{x_{k,1}}{2}\right)\sin\left(\frac{x_{k,0}}{2}\right) \\ \cos\left(\frac{x_{k,1}}{2}\right)\sin\left(\frac{x_{k,0}}{2}\right) & \cos\left(\frac{x_{k,1}}{2}\right)\cos\left(\frac{x_{k,0}}{2}\right) & -\sin\left(\frac{x_{k,1}}{2}\right)\sin\left(\frac{x_{k,0}}{2}\right) & -\sin\left(\frac{x_{k,1}}{2}\right)\cos\left(\frac{x_{k,0}}{2}\right) \\ \sin\left(\frac{x_{k,1}}{2}\right)\cos\left(\frac{x_{k,0}}{2}\right) & -\sin\left(\frac{x_{k,1}}{2}\right)\sin\left(\frac{x_{k,0}}{2}\right) & \cos\left(\frac{x_{k,1}}{2}\right)\cos\left(\frac{x_{k,0}}{2}\right) & -\cos\left(\frac{x_{k,1}}{2}\right)\sin\left(\frac{x_{k,0}}{2}\right) \\ \sin\left(\frac{x_{k,1}}{2}\right)\sin\left(\frac{x_{k,0}}{2}\right) & \sin\left(\frac{x_{k,1}}{2}\right)\cos\left(\frac{x_{k,0}}{2}\right) & \cos\left(\frac{x_{k,1}}{2}\right)\sin\left(\frac{x_{k,0}}{2}\right) & \cos\left(\frac{x_{k,1}}{2}\right)\cos\left(\frac{x_{k,0}}{2}\right)\end{bmatrix}\begin{bmatrix}1 \\ 1 \\ 1 \\ 1\end{bmatrix}$$

$$= \begin{bmatrix}\frac{1}{2}\left(\cos\left(\frac{x_{k,1}}{2}\right)\cos\left(\frac{x_{k,0}}{2}\right) - \cos\left(\frac{x_{k,1}}{2}\right)\sin\left(\frac{x_{k,0}}{2}\right) - \sin\left(\frac{x_{k,1}}{2}\right)\cos\left(\frac{x_{k,0}}{2}\right) + \sin\left(\frac{x_{k,1}}{2}\right)\sin\left(\frac{x_{k,0}}{2}\right)\right) \\ \frac{1}{2}\left(\cos\left(\frac{x_{k,1}}{2}\right)\sin\left(\frac{x_{k,0}}{2}\right) + \cos\left(\frac{x_{k,1}}{2}\right)\cos\left(\frac{x_{k,0}}{2}\right) - \sin\left(\frac{x_{k,1}}{2}\right)\sin\left(\frac{x_{k,0}}{2}\right) - \sin\left(\frac{x_{k,1}}{2}\right)\cos\left(\frac{x_{k,0}}{2}\right)\right) \\ \frac{1}{2}\left(\sin\left(\frac{x_{k,1}}{2}\right)\cos\left(\frac{x_{k,0}}{2}\right) - \sin\left(\frac{x_{k,1}}{2}\right)\sin\left(\frac{x_{k,0}}{2}\right) + \cos\left(\frac{x_{k,1}}{2}\right)\cos\left(\frac{x_{k,0}}{2}\right) - \cos\left(\frac{x_{k,1}}{2}\right)\sin\left(\frac{x_{k,0}}{2}\right)\right) \\ \frac{1}{2}\left(\sin\left(\frac{x_{k,1}}{2}\right)\sin\left(\frac{x_{k,0}}{2}\right) + \sin\left(\frac{x_{k,1}}{2}\right)\cos\left(\frac{x_{k,0}}{2}\right) + \cos\left(\frac{x_{k,1}}{2}\right)\sin\left(\frac{x_{k,0}}{2}\right) + \cos\left(\frac{x_{k,1}}{2}\right)\cos\left(\frac{x_{k,0}}{2}\right)\right)\end{bmatrix} = \begin{bmatrix}v_{0,0}' \\ v_{0,1}' \\ v_{1,0}' \\ v_{1,1}'\end{bmatrix} \tag{A3}$$

$$\left(R(w_{s,1}) \otimes R(w_{s,0})\right)C(q_0)|q_1 q_0\rangle'$$

$$= \begin{bmatrix}\cos\left(\frac{w_{s,1}}{2}\right)\cos\left(\frac{w_{s,0}}{2}\right) & -\cos\left(\frac{w_{s,1}}{2}\right)\sin\left(\frac{w_{s,0}}{2}\right) & -\sin\left(\frac{w_{s,1}}{2}\right)\cos\left(\frac{w_{s,0}}{2}\right) & \sin\left(\frac{w_{s,1}}{2}\right)\sin\left(\frac{w_{s,0}}{2}\right) \\ \cos\left(\frac{w_{s,1}}{2}\right)\sin\left(\frac{w_{s,0}}{2}\right) & \cos\left(\frac{w_{s,1}}{2}\right)\cos\left(\frac{w_{s,0}}{2}\right) & -\sin\left(\frac{w_{s,1}}{2}\right)\sin\left(\frac{w_{s,0}}{2}\right) & -\sin\left(\frac{w_{s,1}}{2}\right)\cos\left(\frac{w_{s,0}}{2}\right) \\ \sin\left(\frac{w_{s,1}}{2}\right)\cos\left(\frac{w_{s,0}}{2}\right) & -\sin\left(\frac{w_{s,1}}{2}\right)\sin\left(\frac{w_{s,0}}{2}\right) & \cos\left(\frac{w_{s,1}}{2}\right)\cos\left(\frac{w_{s,0}}{2}\right) & -\cos\left(\frac{w_{s,1}}{2}\right)\sin\left(\frac{w_{s,0}}{2}\right) \\ \sin\left(\frac{w_{s,1}}{2}\right)\sin\left(\frac{w_{s,0}}{2}\right) & \sin\left(\frac{w_{s,1}}{2}\right)\cos\left(\frac{w_{s,0}}{2}\right) & \cos\left(\frac{w_{s,1}}{2}\right)\sin\left(\frac{w_{s,0}}{2}\right) & \cos\left(\frac{w_{s,1}}{2}\right)\cos\left(\frac{w_{s,0}}{2}\right)\end{bmatrix}\begin{bmatrix}1 & 0 & 0 & 0 \\ 0 & 0 & 0 & 1 \\ 0 & 0 & 1 & 0 \\ 0 & 1 & 0 & 0\end{bmatrix}|q_1 q_0\rangle'$$

$$= \begin{bmatrix}\cos\left(\frac{w_{s,1}}{2}\right)\cos\left(\frac{w_{s,0}}{2}\right) & \sin\left(\frac{w_{s,1}}{2}\right)\sin\left(\frac{w_{s,0}}{2}\right) & -\sin\left(\frac{w_{s,1}}{2}\right)\cos\left(\frac{w_{s,0}}{2}\right) & -\cos\left(\frac{w_{s,1}}{2}\right)\sin\left(\frac{w_{s,0}}{2}\right) \\ \cos\left(\frac{w_{s,1}}{2}\right)\sin\left(\frac{w_{s,0}}{2}\right) & -\sin\left(\frac{w_{s,1}}{2}\right)\cos\left(\frac{w_{s,0}}{2}\right) & -\sin\left(\frac{w_{s,1}}{2}\right)\sin\left(\frac{w_{s,0}}{2}\right) & \cos\left(\frac{w_{s,1}}{2}\right)\cos\left(\frac{w_{s,0}}{2}\right) \\ \sin\left(\frac{w_{s,1}}{2}\right)\cos\left(\frac{w_{s,0}}{2}\right) & -\cos\left(\frac{w_{s,1}}{2}\right)\sin\left(\frac{w_{s,0}}{2}\right) & \cos\left(\frac{w_{s,1}}{2}\right)\cos\left(\frac{w_{s,0}}{2}\right) & -\sin\left(\frac{w_{s,1}}{2}\right)\sin\left(\frac{w_{s,0}}{2}\right) \\ \sin\left(\frac{w_{s,1}}{2}\right)\sin\left(\frac{w_{s,0}}{2}\right) & \cos\left(\frac{w_{s,1}}{2}\right)\cos\left(\frac{w_{s,0}}{2}\right) & \cos\left(\frac{w_{s,1}}{2}\right)\sin\left(\frac{w_{s,0}}{2}\right) & \sin\left(\frac{w_{s,1}}{2}\right)\cos\left(\frac{w_{s,0}}{2}\right)\end{bmatrix}\begin{bmatrix}v_{0,0}' \\ v_{0,1}' \\ v_{1,0}' \\ v_{1,1}'\end{bmatrix}$$

$$= \begin{bmatrix}\cos\left(\frac{w_{s,1}}{2}\right)\cos\left(\frac{w_{s,0}}{2}\right)v_{0,0}' + \sin\left(\frac{w_{s,1}}{2}\right)\sin\left(\frac{w_{s,0}}{2}\right)v_{0,1}' - \sin\left(\frac{w_{s,1}}{2}\right)\cos\left(\frac{w_{s,0}}{2}\right)v_{1,0}' - \cos\left(\frac{w_{s,1}}{2}\right)\sin\left(\frac{w_{s,0}}{2}\right)v_{1,1}' \\ \cos\left(\frac{w_{s,1}}{2}\right)\sin\left(\frac{w_{s,0}}{2}\right)v_{0,0}' - \sin\left(\frac{w_{s,1}}{2}\right)\cos\left(\frac{w_{s,0}}{2}\right)v_{0,1}' - \sin\left(\frac{w_{s,1}}{2}\right)\sin\left(\frac{w_{s,0}}{2}\right)v_{1,0}' + \cos\left(\frac{w_{s,1}}{2}\right)\cos\left(\frac{w_{s,0}}{2}\right)v_{1,1}' \\ \sin\left(\frac{w_{s,1}}{2}\right)\cos\left(\frac{w_{s,0}}{2}\right)v_{0,0}' - \cos\left(\frac{w_{s,1}}{2}\right)\sin\left(\frac{w_{s,0}}{2}\right)v_{0,1}' + \cos\left(\frac{w_{s,1}}{2}\right)\cos\left(\frac{w_{s,0}}{2}\right)v_{1,0}' - \sin\left(\frac{w_{s,1}}{2}\right)\sin\left(\frac{w_{s,0}}{2}\right)v_{1,1}' \\ \sin\left(\frac{w_{s,1}}{2}\right)\sin\left(\frac{w_{s,0}}{2}\right)v_{0,0}' + \cos\left(\frac{w_{s,1}}{2}\right)\cos\left(\frac{w_{s,0}}{2}\right)v_{0,1}' + \cos\left(\frac{w_{s,1}}{2}\right)\sin\left(\frac{w_{s,0}}{2}\right)v_{1,0}' + \sin\left(\frac{w_{s,1}}{2}\right)\cos\left(\frac{w_{s,0}}{2}\right)v_{1,1}'\end{bmatrix} = \begin{bmatrix}v_{0,0}'' \\ v_{0,1}'' \\ v_{1,0}'' \\ v_{1,1}''\end{bmatrix} = |q_1 q_0\rangle'' \tag{A4}$$

$$\left(\prod_{t=1}^{T}\left(R(w_{s,2(T-t)+1}) \otimes R(w_{s,2(T-t)})\right)C(q_0)\right)|q_1 q_0\rangle'$$

$$= \left(\prod_{t=1}^{T}\begin{bmatrix}\cos\left(\frac{w_{s,2(T-t)+1}}{2}\right)\cos\left(\frac{w_{s,2(T-t)}}{2}\right) & -\cos\left(\frac{w_{s,2(T-t)+1}}{2}\right)\sin\left(\frac{w_{s,2(T-t)}}{2}\right) & -\sin\left(\frac{w_{s,2(T-t)+1}}{2}\right)\cos\left(\frac{w_{s,2(T-t)}}{2}\right) & \sin\left(\frac{w_{s,2(T-t)+1}}{2}\right)\sin\left(\frac{w_{s,2(T-t)}}{2}\right) \\ \cos\left(\frac{w_{s,2(T-t)+1}}{2}\right)\sin\left(\frac{w_{s,2(T-t)}}{2}\right) & \cos\left(\frac{w_{s,2(T-t)+1}}{2}\right)\cos\left(\frac{w_{s,2(T-t)}}{2}\right) & -\sin\left(\frac{w_{s,2(T-t)+1}}{2}\right)\sin\left(\frac{w_{s,2(T-t)}}{2}\right) & -\sin\left(\frac{w_{s,2(T-t)+1}}{2}\right)\cos\left(\frac{w_{s,2(T-t)}}{2}\right) \\ \sin\left(\frac{w_{s,2(T-t)+1}}{2}\right)\cos\left(\frac{w_{s,2(T-t)}}{2}\right) & -\sin\left(\frac{w_{s,2(T-t)+1}}{2}\right)\sin\left(\frac{w_{s,2(T-t)}}{2}\right) & \cos\left(\frac{w_{s,2(T-t)+1}}{2}\right)\cos\left(\frac{w_{s,2(T-t)}}{2}\right) & -\cos\left(\frac{w_{s,2(T-t)+1}}{2}\right)\sin\left(\frac{w_{s,2(T-t)}}{2}\right) \\ \sin\left(\frac{w_{s,2(T-t)+1}}{2}\right)\sin\left(\frac{w_{s,2(T-t)}}{2}\right) & \sin\left(\frac{w_{s,2(T-t)+1}}{2}\right)\cos\left(\frac{w_{s,2(T-t)}}{2}\right) & \cos\left(\frac{w_{s,2(T-t)+1}}{2}\right)\sin\left(\frac{w_{s,2(T-t)}}{2}\right) & \cos\left(\frac{w_{s,2(T-t)+1}}{2}\right)\cos\left(\frac{w_{s,2(T-t)}}{2}\right)\end{bmatrix}\begin{bmatrix}1 & 0 & 0 & 0 \\ 0 & 0 & 0 & 1 \\ 0 & 0 & 1 & 0 \\ 0 & 1 & 0 & 0\end{bmatrix}\right)|q_1 q_0\rangle'$$

$$= \left(\prod_{t=1}^{T}\begin{bmatrix}\cos\left(\frac{w_{s,2(T-t)+1}}{2}\right)\cos\left(\frac{w_{s,2(T-t)}}{2}\right) & \sin\left(\frac{w_{s,2(T-t)+1}}{2}\right)\sin\left(\frac{w_{s,2(T-t)}}{2}\right) & -\sin\left(\frac{w_{s,2(T-t)+1}}{2}\right)\cos\left(\frac{w_{s,2(T-t)}}{2}\right) & -\cos\left(\frac{w_{s,2(T-t)+1}}{2}\right)\sin\left(\frac{w_{s,2(T-t)}}{2}\right) \\ \cos\left(\frac{w_{s,2(T-t)+1}}{2}\right)\sin\left(\frac{w_{s,2(T-t)}}{2}\right) & -\sin\left(\frac{w_{s,2(T-t)+1}}{2}\right)\cos\left(\frac{w_{s,2(T-t)}}{2}\right) & -\sin\left(\frac{w_{s,2(T-t)+1}}{2}\right)\sin\left(\frac{w_{s,2(T-t)}}{2}\right) & \cos\left(\frac{w_{s,2(T-t)+1}}{2}\right)\cos\left(\frac{w_{s,2(T-t)}}{2}\right) \\ \sin\left(\frac{w_{s,2(T-t)+1}}{2}\right)\cos\left(\frac{w_{s,2(T-t)}}{2}\right) & -\cos\left(\frac{w_{s,2(T-t)+1}}{2}\right)\sin\left(\frac{w_{s,2(T-t)}}{2}\right) & \cos\left(\frac{w_{s,2(T-t)+1}}{2}\right)\cos\left(\frac{w_{s,2(T-t)}}{2}\right) & -\sin\left(\frac{w_{s,2(T-t)+1}}{2}\right)\sin\left(\frac{w_{s,2(T-t)}}{2}\right) \\ \sin\left(\frac{w_{s,2(T-t)+1}}{2}\right)\sin\left(\frac{w_{s,2(T-t)}}{2}\right) & \cos\left(\frac{w_{s,2(T-t)+1}}{2}\right)\cos\left(\frac{w_{s,2(T-t)}}{2}\right) & \cos\left(\frac{w_{s,2(T-t)+1}}{2}\right)\sin\left(\frac{w_{s,2(T-t)}}{2}\right) & \sin\left(\frac{w_{s,2(T-t)+1}}{2}\right)\cos\left(\frac{w_{s,2(T-t)}}{2}\right)\end{bmatrix}\right)|q_1 q_0\rangle' = \begin{bmatrix}v_{0,0}'' \\ v_{0,1}'' \\ v_{1,0}'' \\ v_{1,1}''\end{bmatrix} = |q_1 q_0\rangle'' \tag{A5}$$